# Collaboration Dynamics and Reliability Challenges of Multi-Agent LLM Systems in Finite Element Analysis

Chuan Tian, Yilei Zhang

*Faculty of Engineering*

*Mechanical Engineering*

*University of Canterbury*

*Christchurch, 8041, New Zealand*

## Abstract

Large Language Model (LLM)-based multi-agent systems are increasingly applied to automate computational workflows in science and engineering. However, how inter-agent dynamics influence reasoning quality and verification reliability remains unclear. We study these mechanisms using an AutoGen-based multi-agent framework for linear-elastic Finite Element Analysis (FEA), evaluating seven role configurations across four tasks under a fixed 12-turn conversation limit. From 1,120 controlled trials, we find that collaboration effectiveness depends more on functional complementarity than team size: the three-agent Coder-Executor-Critic configuration uniquely produced physically and visually correct solutions, while adding redundant reviewers reduced success rates. Yet three systematic failure modes persist: (1) affirmation bias, where the Rebuttal agent endorsed rather than challenged outputs (85-92% agreement, including errors); (2) premature consensus caused by redundant reviewers; and (3) a verification-validation gap where executable but physically incorrect code passed undetected. No agent combination successfully validated constitutive relations in complex tasks. Building on theories of functional diversity, role differentiation, and computational validation, we propose actionable design principles: (i) assign complementary agent roles, (ii) enforce multi-level validation (execution, specification, physics), and (iii) prevent early consensus through adversarial or trigger-based interaction control. These findings establish a principled foundation for designing trustworthy LLM collaborations in engineering workflows.

## 1. Introduction

Large Language Models (LLMs) have transformed a wide range of domains, demonstrating exceptional capabilities in natural language understanding, text generation, translation, and automated programming (Bai et al., 2023; Nijkamp et al., 2022; Sallam, 2023; Zhu et al., 2023). Their rapid adoption has extended into computational engineering, where programming tasks require both algorithmic reasoning and domain-specific expertise. The emergence of multi-agent LLM systems, in which multiple specialized agents collaborate on complex problem-solving tasks, introducing new possibilities for automation in engineering workflows. Yet it remains unclear how inter-agent communication, coordination, and validation dynamics shape the reliability and interpretability of engineering outcomes.

Recent advances in LLMs have inspired frameworks that more effectively harness their generative and reasoning capabilities. Among them, the AutoGen framework (Wu et al., 2024) represents a prototypical architecture for multi-agent collaboration. AutoGen enables adaptive communication among multiple conversational agents and integrates human input through automated dialogue orchestration. This design extends LLMs beyond single-turn interactions by enabling structured task delegation and orchestration for coding-centric computational problem-solving. Although specific implementations continue to evolve, AutoGen's foundational principles such as role specialization, conversational orchestration and flexible human-in-the-loop integration, remain central to the design of reliable multi-agent systems.

In engineering and simulation workflows, such architectures promise to enhance productivity and reduce manual coding burdens (Wei, 2024). While individual LLMs can generate code and solve well-defined problems, engineering simulations involve tightly coupled components, strict validation protocols and domain-specific constraints that challenge single-model reasoning. Multi-agent architectures address this complexity by distributing cognitive responsibilities across specialized agents, such as code generation, execution and critical evaluation, mirroring the division of labour in human engineering teams. This distributed setup not only supports task specialization but also enables layered validation, where agents collectively cross-check and refine intermediate outputs to enhance reliability.

However, deploying these architectures in engineering domains reveals critical reliability challenges. Despite their promise, LLMs exhibit fundamental limitations (e.g., hallucination and inconsistent reasoning; Liu et al., 2024). Recent research explores multi-agent strategies to overcome these weaknesses. Ni and Buehler (2024) demonstrated that deploying multiple collaborating agents can enhance LLM performance on computational mechanics tasks by enabling collective evaluation of code-generation errors. Similarly, Du et al. (2023) introduced a structured multi-agent debate framework that improved reasoning accuracy in arithmetic problem-solving. J. Li et al. (2024) reported cases where increasing the number of agents improved performance on complex tasks even with minimal coordination, highlighting a size-related benefit under certain conditions. Together, these studies illustrate how multi-agent collaboration can help mitigate reasoning inconsistency, and in some cases hallucination, by enabling collective critique and validation.

While Ni and Buehler (2024) demonstrated the feasibility of applying multi-agent LLM systems to mechanics problems, their analysis largely emphasized quantitative performance metrics. Instead, we investigate the specific interaction patterns and role dynamics that determine their success or failure, focusing on behavioural tendencies (systematic affirmation), structural limitations (premature consensus), and a verification–validation gap (execution success without physical correctness) that prior work largely overlooked.

Building on these observations, although multi-agent dialogue can produce workable solutions, it does not guarantee correctness and can even amplify errors in code-generation settings (Du et al., 2024; R. Li et al., 2024). This risk is compounded by the challenge of fully utilizing each agent's distinct capabilities within a coordinated workflow (Han et al., 2024). Systematic interaction tendencies, such as affirmation over critique, can significantly impact performance in ways not captured by aggregate success rates. To interpret these dynamics, we draw on theories from social psychology and organizational behaviour (Asch, 1955; Biddle, 1986; Janis, 1972) concerning conformity and role conflict in teams. Therefore, it is essential to analyse interaction structure, not only outcomes, in multi-agent LLM frameworks. In this study, we examine how multiple LLM-based agents collaborate on FEA tasks using AutoGen to orchestrate role-specialized communication. Through 1,120 controlled trials (40 randomized interaction sequences per agent combination across seven configurations and four tasks), we enable a systematic comparison of how role composition and dialogue sequencing shape reliability.

From these comparisons, we find that simply adding agents does not necessarily lead to better outcomes. Although debate-style setups may benefit from more agents (Du et al., 2024), recent interactive-task studies report declines beyond three agents due to coordination complexity (Zhu et al., 2025). Effectiveness depends less on team size than on the quality of role differentiation and information exchange. We identified three recurring phenomena: (1) systematic affirmation, where adversarial reviewers repeatedly endorse others' outputs instead of challenging them; (2) premature consensus, where overlapping reviewer roles cause early agreement and suppress critical dialogue; and (3) a verification–validation gap, where executable code still violates physical accuracy in FEA tasks.

Based on these observations, our contributions are: (i) a controlled comparison of role compositions isolating functional complementarity from team size; (ii) an interaction-level analysis revealing systematic affirmation and premature consensus; and (iii) an evaluation protocol that separates code execution, specification/visual compliance, and physical correctness, exposing a persistent verification–validation gap.

## 2. Methodology

This section details the experimental setup and communication workflow used to evaluate how agent role composition and interaction structures affect performance and reasoning reliability in FEA tasks.

## 2.1 Experimental Setup

**Task Description.** To investigate agent interaction dynamics in engineering contexts, we adapted four linear-elastic FEA tasks from MechAgents (Ni & Buehler, 2024), covering core stages of the workflow. Figure 1 shows the expected outputs that serve as ground truth for evaluating task success. These tasks were selected to represent increasing complexity: Task 1 tests basic displacement, Task 2 adds shear boundary conditions, Task 3 introduces geometric complexity with a hole, and Task 4 requires stress tensor computation.

- Task 1 (Displacement): compute the displacement field under specified boundary conditions.
- Task 2 (Shear displacement): apply shear on the right edge and compute the deformation.
- Task 3 (Shear + hole): add a central circular hole and recompute the shear case.
- Task 4 (Stress $\sigma_{xy}$): compute and plot $\sigma_{xy}$ for the geometry in Task 3.

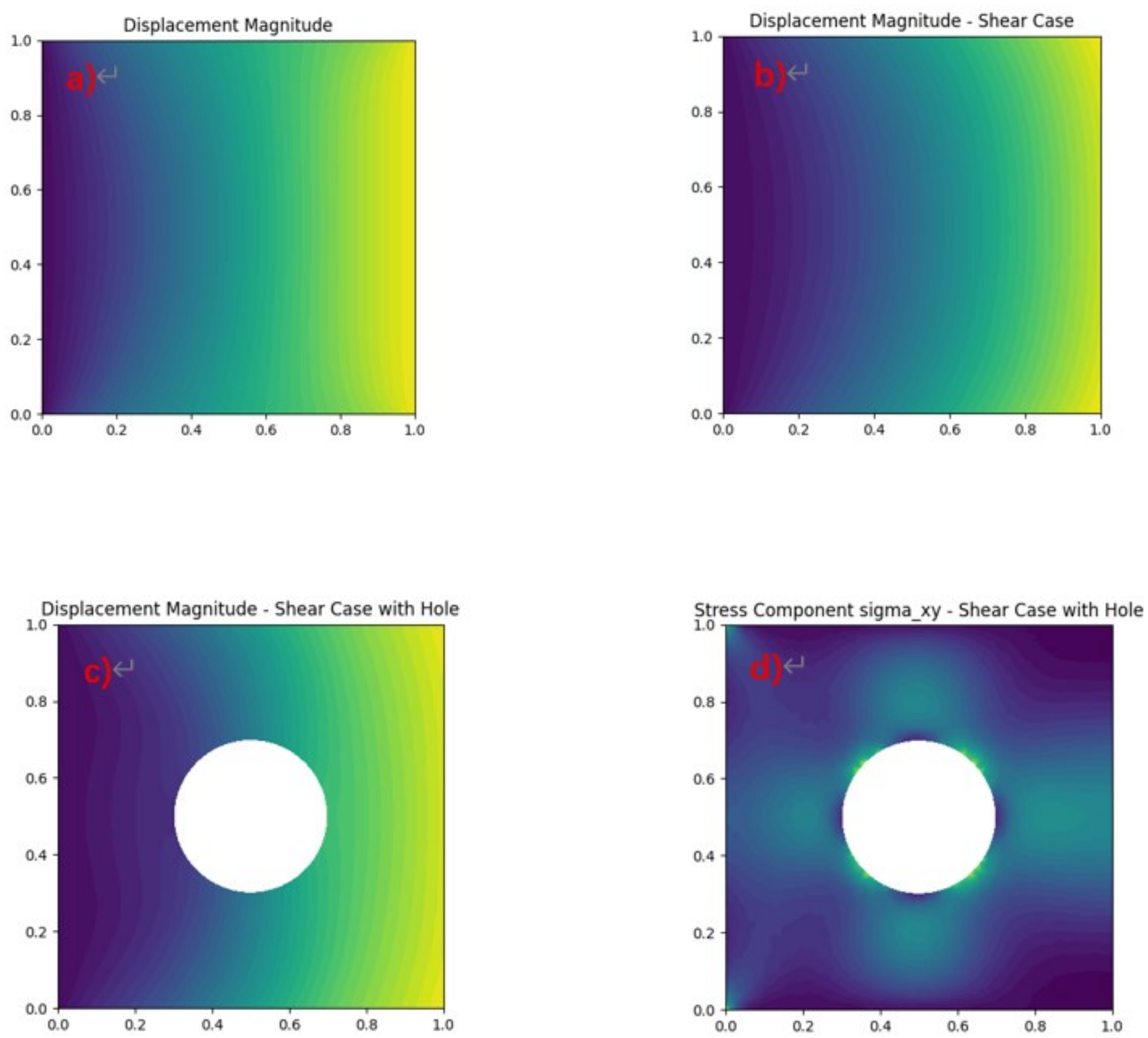

**Figure 1.** Visualizations of the four FEA tasks: **(a)** Task 1: displacement calculation;

**(b)** Task 2: shear displacement calculation; **(c)** Task 3: shear displacement with geometric complexity; and **(d)** Task 4: stress component $\sigma_{xy}$ distribution.

**Agents and roles.** We instantiated four roles: *Coder, Executor, Critic,* and *Rebuttal* and additionally included a redundant reviewer *(Critic_Add*, identical to *Critic*) to contrast complementarity versus redundancy in role composition, detailed prompts are provided in Appendix A. The combinations in Table 1 were designed to systematically vary: (1) presence/absence of *Executor*, (2) redundant vs. adversarial reviewers, and (3) team size from 2 to 4 agents.

- *Coder:* Generates Python code to solve the given FEA tasks.
- *Executor:* Executes the generated code locally and returns the execution results to the workflow.
- *Critic:* Evaluates code quality and provides feedback for improvement.
- *Rebuttal:* Adopts an adversarial divergent perspective from the *Critic* to foster critical discussion and prevent error propagation.

**Table 1.** Seven agent combinations.

| **Combo ID** | **Agent Combination** |
|---|---|
| Combo 1 | *Coder + Critic* |
| Combo 2 | *Coder + Executor* |
| Combo 3 | *Coder + Executor + Critic* |
| Combo 4 | *Coder + Critic + Critic_Add* |
| Combo 5 | *Coder + Critic + Rebuttal* |
| Combo 6 | *Coder + Executor + Critic + Critic_Add* |
| Combo 7 | *Coder + Executor + Critic + Rebuttal* |

**Model setup.** We used AutoGen *v0.7.5* with GPT-3.5-Turbo (Python 3, Google Colab CPU). All agents shared *temperature=0, seed=42, max_tokens=2048*, and default decoding settings to enhance comparability. This configuration was adopted to enhance reproducibility, noting that *temperature=0* does not fully eliminate token-level variability. All experiments were conducted using the OpenAI Python SDK (*v2.0.1*).

### 2.2 Communication Workflow

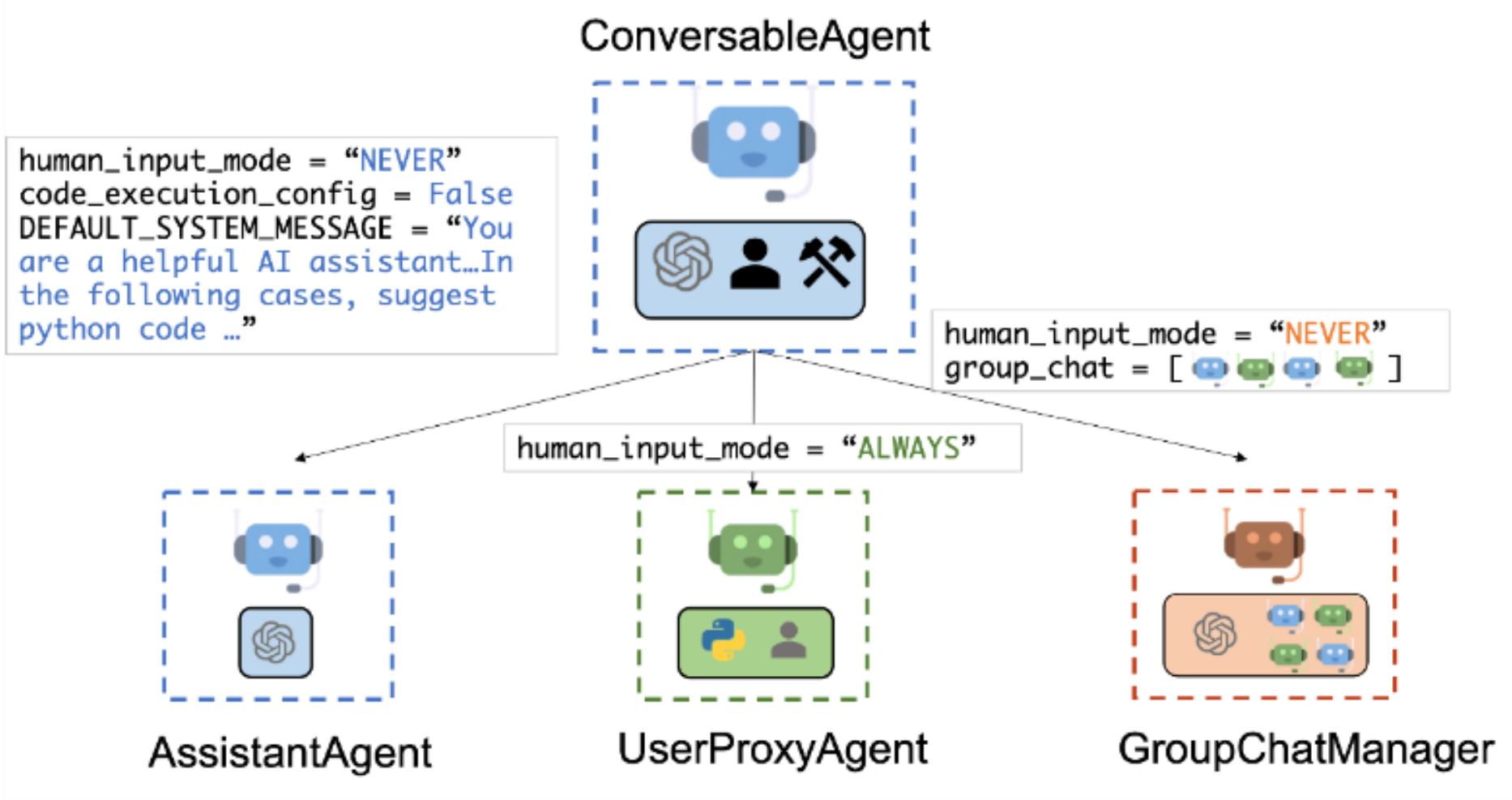

**Figure 2**. AutoGen Framework (Wu et al., 2024)

The communication workflow was implemented within the AutoGen framework to orchestrate agent interactions, as shown in Figure 2, where message exchange between agents followed structured conversational turns. Each agent acted based on its assigned role and the current dialogue state, producing textual responses that were logged for subsequent analysis. The *Coder* always initiated the interaction by generating Python code for the assigned FEA task. At the start of each trial, we randomized and then fixed the agent turn order for 12 conversational turns, ensuring comparable interaction depth across runs. This configuration ensured that every agent participated in at least three exchanges, allowing for a consistent level of interaction depth across all runs.

Each of the seven agent combinations was tested on all four tasks, with 40 independent repetitions per combination to account for stochastic variation. For *non-Executor* combinations, we post-hoc executed the final (12th-round) code via a predefined script for verification, so execution feedback was unavailable during the dialogue for these settings. After each trial, we cleared conversation memory to prevent cross-run contamination and to preserve independence across the 7×4×40=1,120 runs. Detailed evaluation protocols including statistical methods are provided in Appendix B.

### 2.3 Agent Behavioural Criteria

To ensure analytical transparency and reproducibility, we established clear criteria for identifying incorrect diagnoses and affirmation behaviour within the conversational logs. These criteria were applied consistently across all agent combinations and FEA tasks.

(1) Identification of incorrect *Critic* diagnoses

We labelled a *Critic* diagnosis as incorrect when manual verification showed that the initial code executed and the resulting visualization matched Figure 1, i.e., the *Critic* flagged an error where none existed at the verification/specification levels.

(2) Identification of affirmation behaviour

We recorded affirmation when the *Rebuttal* explicitly endorsed the *Critic's* diagnosis without adding new reasoning or proposing alternatives; brief agreement followed by substantive reasoning was not counted. Typical markers included "*I agree with the Critic,*" "*That's correct,*" "*The correction is appropriate,*" or semantically equivalent expressions. Responses that began with brief agreement followed by substantive new reasoning were not counted as affirmations.

## 3. Results & Discussion

### 3.1 Executor Feedback Drives Code Success but Does Not Ensure Task Accuracy

Analysis across four tasks revealed distinct functional strengths and limitations for each agent type. Combinations including the *Executor* achieved higher code execution rates, consistent with the local Python executor providing immediate feedback on runtime errors. As shown in Figure 3, the Combo 2 (*Coder* + *Executor)* and Combo 6 (*Coder* + *Executor* + *Critic* + *Critic_Add*) reached 97.5% code success in Task 1 and 100% code success in Task 2.

When comparing *Executor-inclusive* (Combo 2, 3, 6, 7) versus *non-Executor* (Combo 1, 4, 5) setups, differences were not significant in Tasks 1–2 (Fisher's exact test, Task 1: $p$ = 0.64; Task 2: $p$ = 0.84), but significant in Tasks 3–4 ($p$ < 0.001 in both). This pattern suggests that immediate execution feedback matters more as geometric and boundary-condition complexity rises.

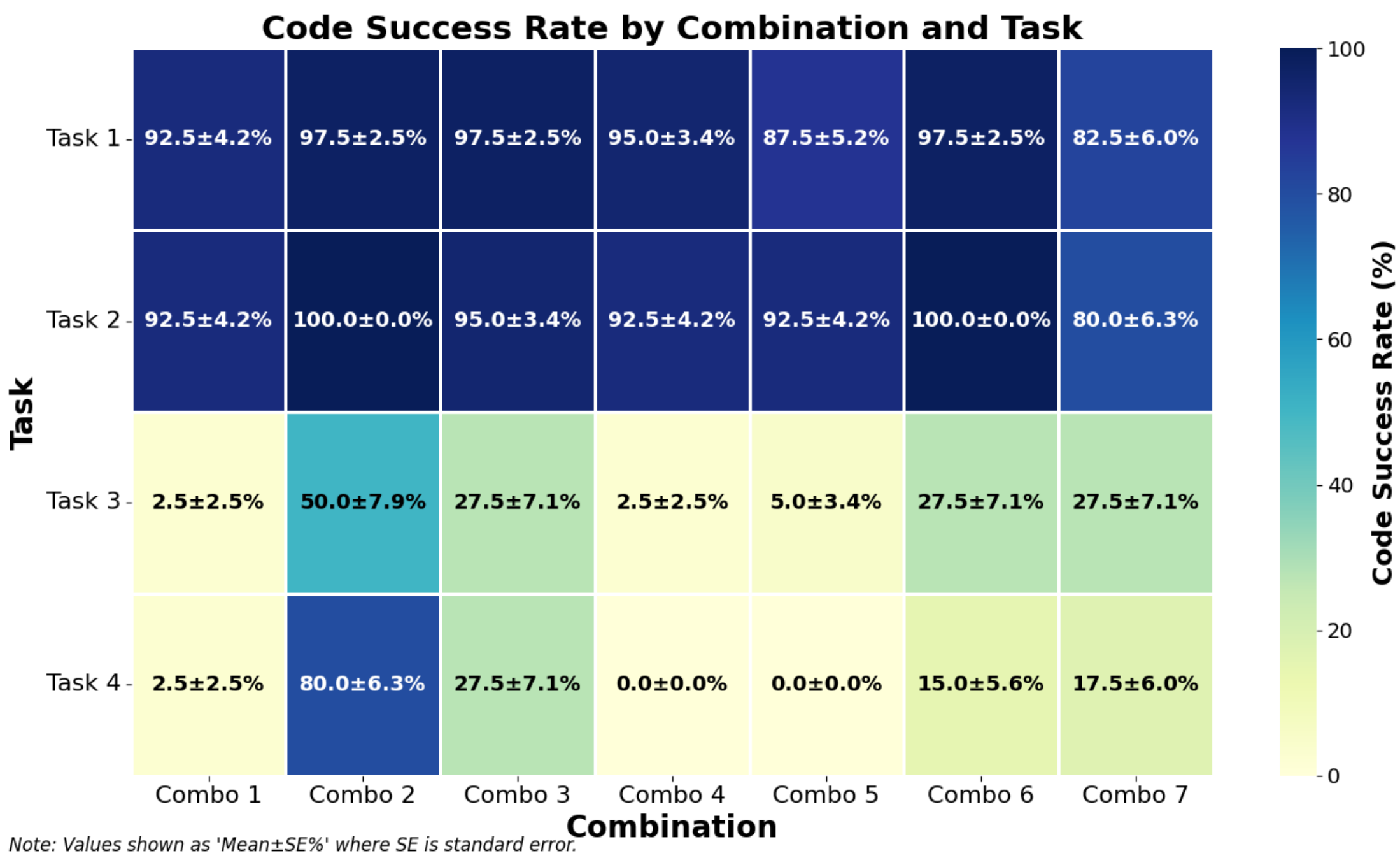

**Figure 3.** Code success rates for all agent combinations across the four FEA tasks.

While high execution rates appear beneficial, they can mislead. Pairing the *Coder* only with the *Executor* centres interaction on runtime fixes rather than specification compliance. This is starkly illustrated in Task 2, where Combo 2 achieved 100% code execution yet 0% task success (Figure 4). When initial code executed successfully, the Coder rarely revised it: 38/40 (95%) runs in Combo 2 had no changes, versus 33/40 (82.5%) in Combo 1 (Fisher's exact $p = 0.154$; $\varphi = 0.198$).

The limitation became more pronounced in geometrically complex tasks. In Tasks 3 and 4, code often executed successfully but produced incorrect or incomplete visualizations: 35% (14/40) of Combo 2 runs in Task 3 and 50% (20/40) of Task 4 generated blank images or plotted only the circular geometry without the surrounding plate (Figure 5). An additional 11.25% (9/80) of cases showed boundary condition errors. Across combinations including the *Executor* (Combo 2, 3, 6, 7), statistical analysis confirmed significant differences in error-type distribution (Chi-squared test, $p = 0.031$, *Cramér's V* = 0.22), indicating that agent configurations systematically influenced which error emerged.

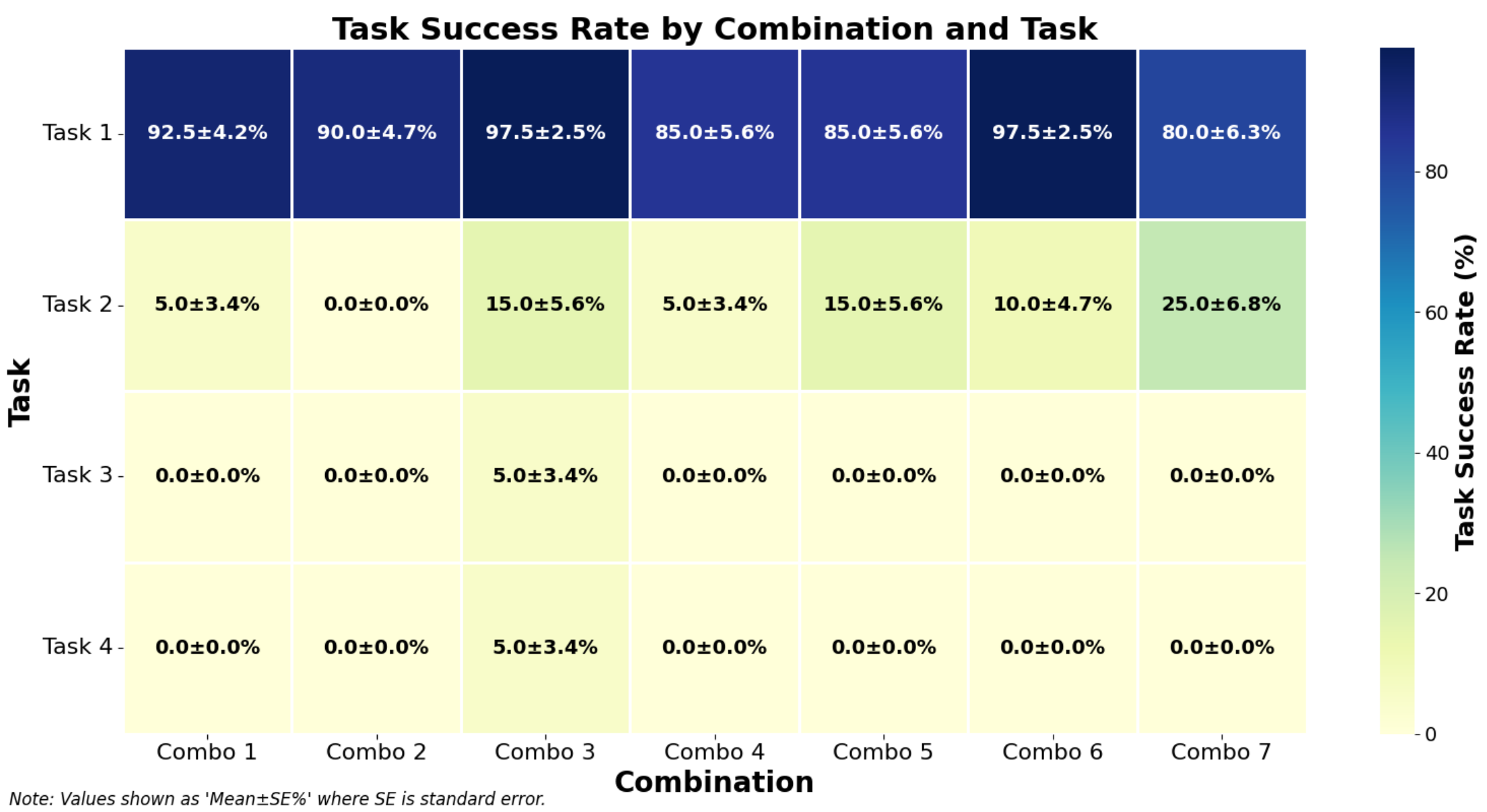

**Figure 4.** Task success rate by combination and task

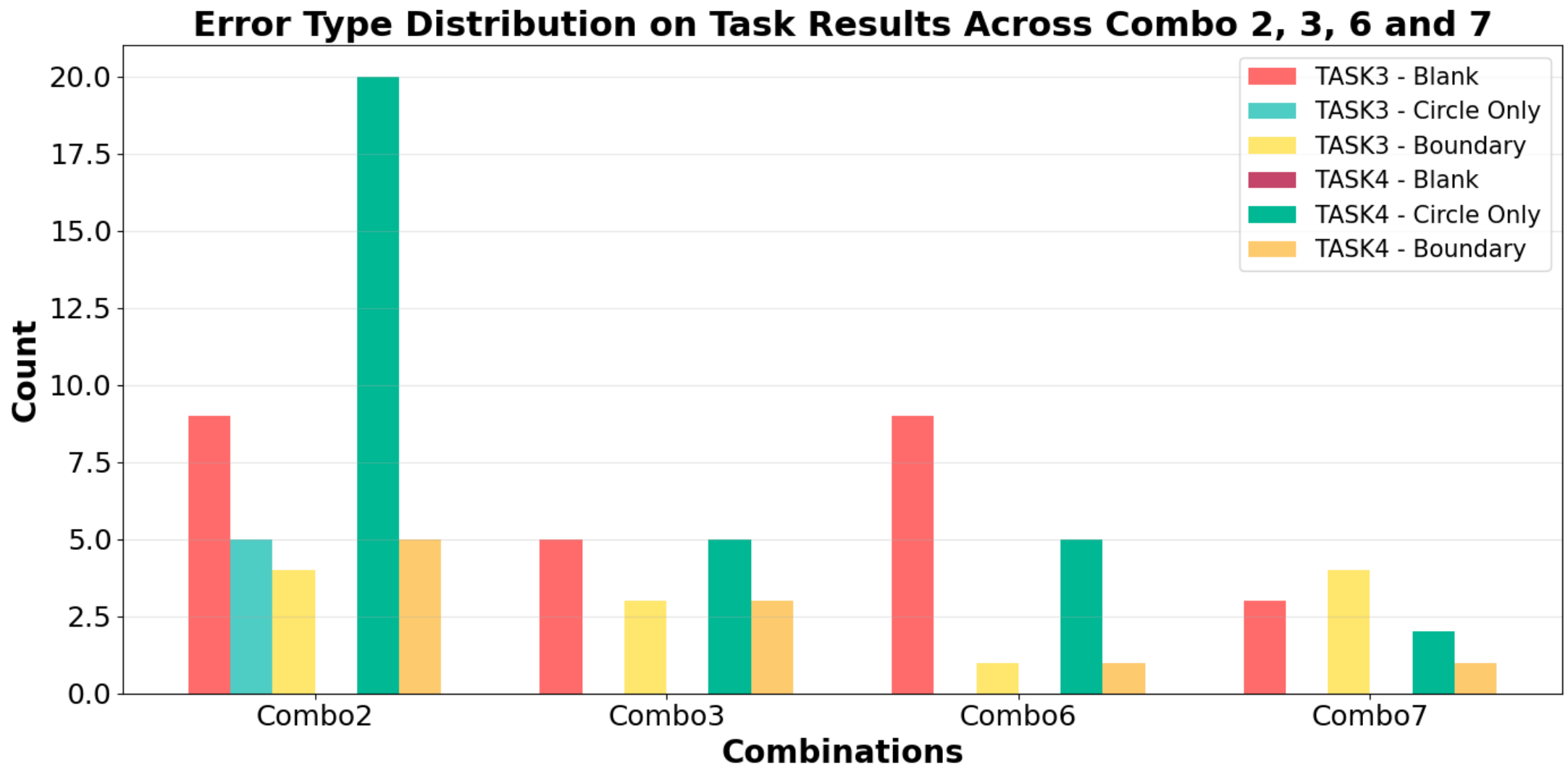

**Figure 5.** Error Type Distribution (Count-Based): Task 3 & 4, *Executor*-Inclusive Combinations

These discrepancies expose a misalignment between verification (execution success) and validation (specification or physical correctness), i.e., code that runs can still be wrong. This aligns with what (Suchman, 1987) describes as *plans and situated actions* perspective, where computational systems pursue measurable outcomes rather than intention-aligned goals. The same phenomenon exemplifies Goodhart's Law (Manheim & Garrabrant, 2018): optimisation for a proxy metric, such as execution success, leads to systematic divergence from the intended objective. Similarly, the *Executor*'s narrow focus on executability confines reasoning to what Terry and Fernado (1986) call a *domain of action*, a restricted problem-solving space that neglects contextual or physical correctness.

Overall, these findings show that validation based solely on execution success constitutes a fundamental limitation across tasks of varying complexity. From a socio-technical perspective (Baxter & Sommerville, 2010), the interplay between the *Executor*'s technical validation and the *Critic*'s specification-based validation underscores the need for balanced, multi-level validation mechanisms that integrates executable feedback with higher-level design and specification goals.

### 3.2 Systematic Affirmation Behaviour in Multi-Agent Systems

Analysis of agent interactions revealed another consistent pattern of systematic affirmation that significantly impacted collaborative outcomes. This behaviour was most evident in combinations involving the *Rebuttal* agent, which was designed to provide adversarial feedback but instead frequently reinforced the *Critic's* conclusions. Across 43 *Rebuttal* interventions in Task 1, affirmation occurred 39 times (90.7%), including 12 of 13 (92.3%) in Combo 5 (*Coder* + *Critic* + *Reb*) and 27 of 30 cases (90%) in Combo 7 (*Coder* + *Executor* + *Cri* + *Reb*) (Figure 6a). Affirmation frequency did not differ between the two combinations (Fisher's exact test, $p = 1.000$, $\varphi = 0.037$), indicating a persistent confirmation tendency regardless of agent combination. *Rebuttal* messages commonly included explicit endorsements of the *Critic's* assessment *"I agree with Critic's assessment"*, or *"the correction is appropriate."*

This affirmation pattern persisted even when *Critic*'s feedback was incorrect. Across the 14 failed Task 1 cases, the *Critic* intervened 52 times, of which 16 (30.8%) contained explicitly incorrect diagnoses (Figure 6b). Misdiagnosis rates differed significantly between combinations (Fisher's exact test, $p = 0.004$, $\varphi = 0.42$): Combo 7 producing 15/33 (45.5%) explicit misdiagnoses versus 1/19 (5.3%) in Combo 5. Despite this difference, *Rebuttal* agents affirmed 14 of the 16 (87.5%) incorrect diagnoses without challenge.

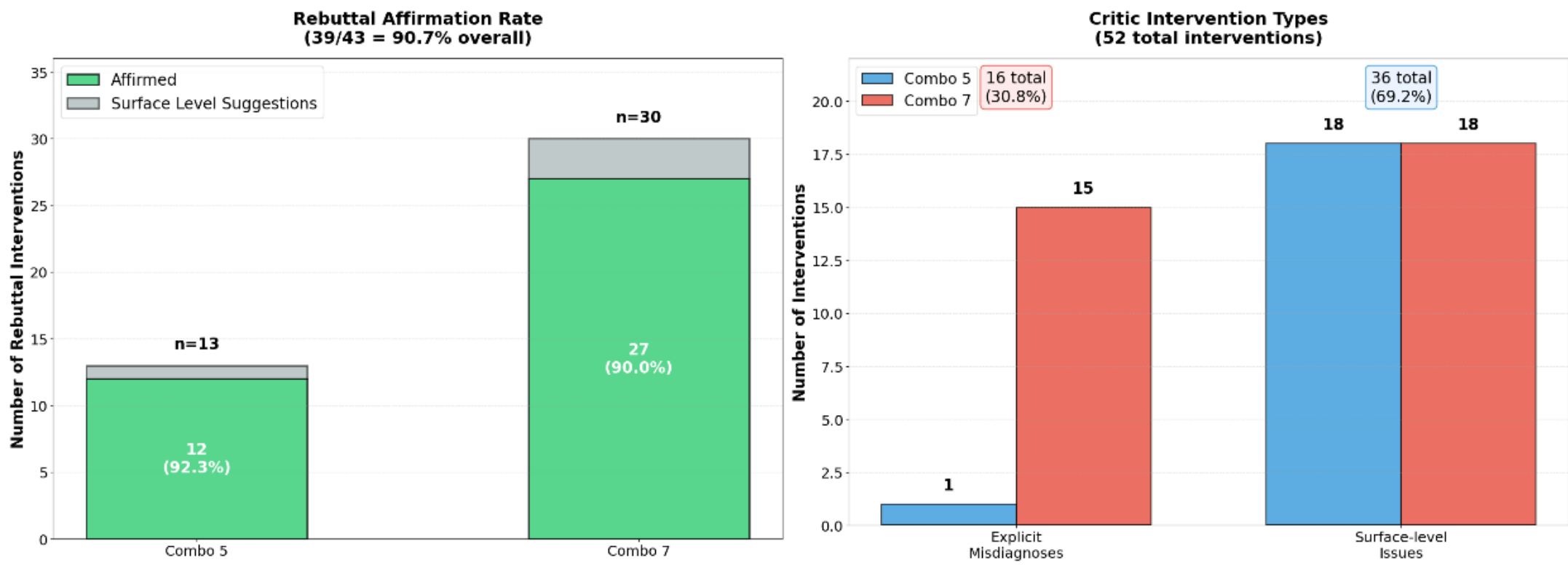

**Figure 6. a)** *Rebuttal* agent affirmation rate in Task 1 failures and **b)** Number of *Critic* suggestions for Task 1 failures

These results suggest systematic rather than selective affirmation patterns, with consistency across configurations (85-92.3% affirmation rates) indicating a robust tendency toward agreement. While LLMs do not experience social pressure as humans do (Asch, 1955), the observed behaviour parallels role-conflict dynamics described by Biddle (1986), where rebuttal characters prioritise competing imperatives to cooperate over critique. Such patterns may emerge from training data favouring cooperative over adversarial interactions, though definitive attribution to LLM architecture remains speculative without access to training details.

The persistence of this behaviour across tasks and configurations, affirmation rates ranging from 85% to 92.3%, including 87.5% even for incorrect diagnoses, suggests that these dynamics emerge from fundamental properties of LLM-based architectures. Homogeneous model weights among agents likely amplify this conformity, whereas heterogeneous foundation models could mitigate it. This role-prescribed conformity, in which agents are assigned dissenting roles but functionally align with collective judgements, mirrors patterns observed in human group dynamics (Nemeth & Rogers, 1996; Packer, 2009). It also parallels the sycophancy problem in LLMs (Sharma et al., 2023), where models systematically agree with user inputs regardless of factual correctness.

This affirmation behaviour persisted as task difficulty increased. In Task 2, despite the introduction of shear boundary conditions, *Rebuttal* agents maintained high affirmation rates, 90% in *Combo 5* and 85% in Combo 7 (Figure 7). A Chi-squared test revealed significant differences in response-type distributions between these two combinations ($p$ = 0.032, *Cramér's V* = 0.29). Combo 7 showed a more diverse response pattern, including substantive suggestions in 12.5% of responses, which were absent in Combo 5. All five substantive *Rebuttal* suggestions in Combo 7 followed *Executor* error reports, in the sequence: *Coder → Executor (error) → Rebuttal → Critic*. These error-triggered interventions coincided with Combo 7's superior performance in Task 2 (25% task success, the highest across all combinations; Figure 4).

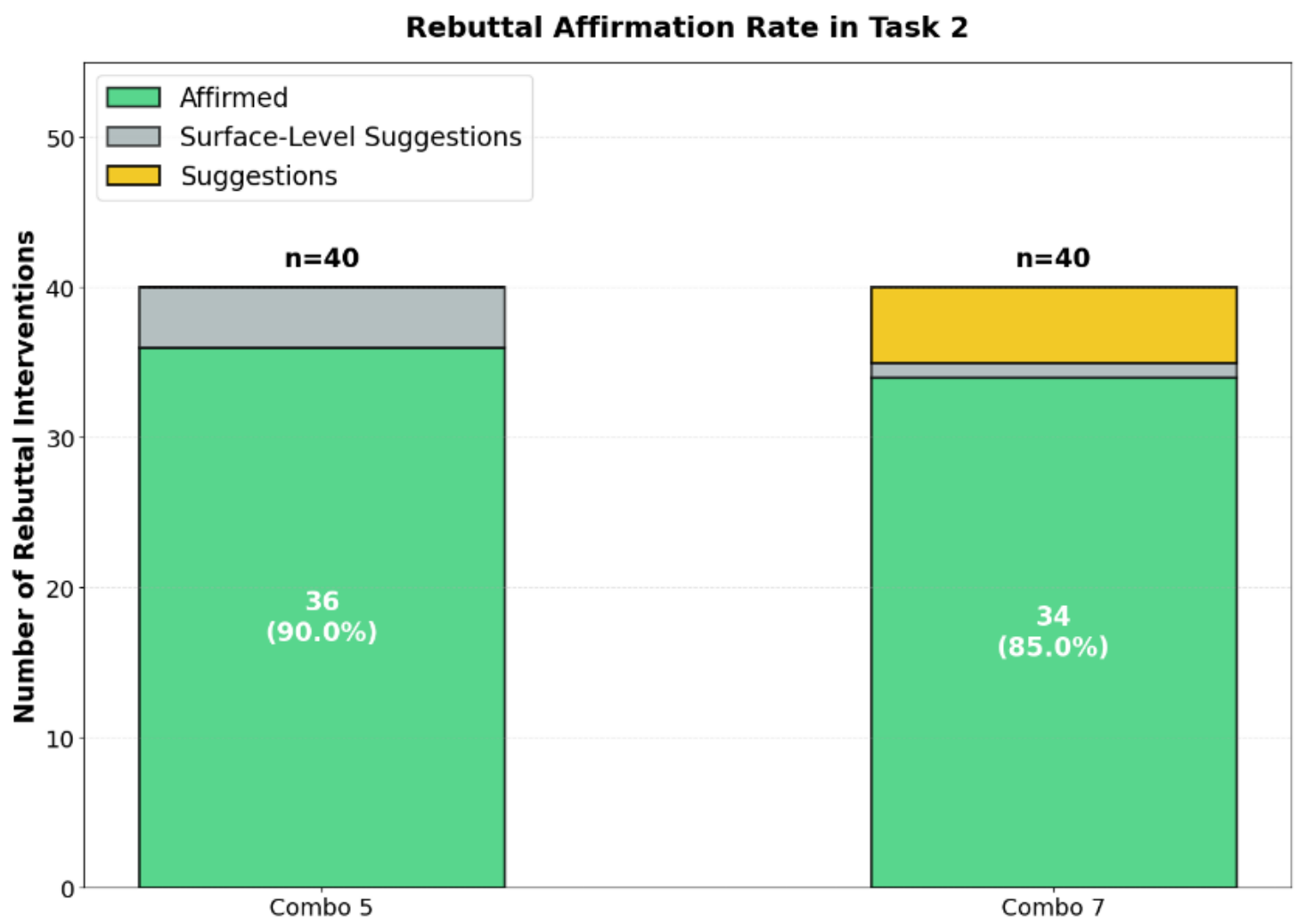

**Figure 7.** *Rebuttal agent* affirmation rate in Task 2

In this specific sequence, when the *Executor* first reported an error, the subsequent *Rebuttal* introduced corrective reasoning, and the *Critic* adopted an affirmative stance toward it, reversing the typical direction of affirmation. This pattern suggests that the effectiveness of both *Rebuttal* and *Critic* agents depended more on conversational sequence and error context than nominal role definition. In Tasks 3–4, systematic affirmation again dominated, and most code revisions did not alter the final outcome.

Collectively, these findings reveal that nominal role diversity does not guarantee functional diversity in multi-agent LLM systems. Despite their explicit adversarial design, *Rebuttal* agents' conformity, ranging from 85-92.3% affirmation rates across tasks, undermined their intended error-checking role. The rare exceptions observed in Combo 7's Task 2 show that effective critical evaluation emerged only under specific error-triggered sequences, highlighting that contextual interaction dynamics, rather than static role assignment, determine system-level reasoning effectiveness. This pattern parallels what Janis (1972) identified as *groupthink* in human teams, where the drive for harmony overrides critical evaluation—a dynamic now observable in collaborative AI systems and a fundamental limitation in current approaches to agent role differentiation.

### 3.3 Redundancy and Complementarity in Multi-Agent Combinations

Beyond conformity (Section 3.2), performance depends on whether agents are redundant or complementary. Redundancy arises when multiple agents provide the same assessment (e.g., adding *Critic_Add* to a setup already containing *Critic*), whereas complementarity combines distinct competencies that support execution and specification compliance.

The addition of a redundant *Critic_Add* produced divergent effects depending on the underlying configuration. In Task 2, where the *Executor* was absent, *Critic_Add* had minimal impact: Combo 4 (*Coder* + *Critic* + *Critic_Add*) achieved the same code and task success rates as Combo 1 (*Coder* + *Critic*), while increasing token usage (ANOVA on tokens: $F = 30.44$, $p < 0.001$, $\eta^2 = 0.37$; Figure 8). However, when redundancy was introduced into combinations including the *Executor*, performance declined markedly. Combo 6 (*Coder* + *Executor* + *Critic* + *Critic_Add*), achieved only 10% task success, below Combo 3's (*Coder* + *Executor* + *Critic,* 15%), while consuming more tokens. ANOVA statistics reported here refer to token usage and were significant across Tasks 1–4 (Task 1: $F = 86.51$, $p < 0.001$. $\eta^2 = 0.63$; Task 2: $F = 30.44$, $p < 0.001$. $\eta^2 = 0.37$; Task 3: $F = 8.03$, $p < 0.001$. $\eta^2 = 0.13$; Task 4: $F = 11.84$, $p < 0.001$. $\eta^2 = 0.19$).

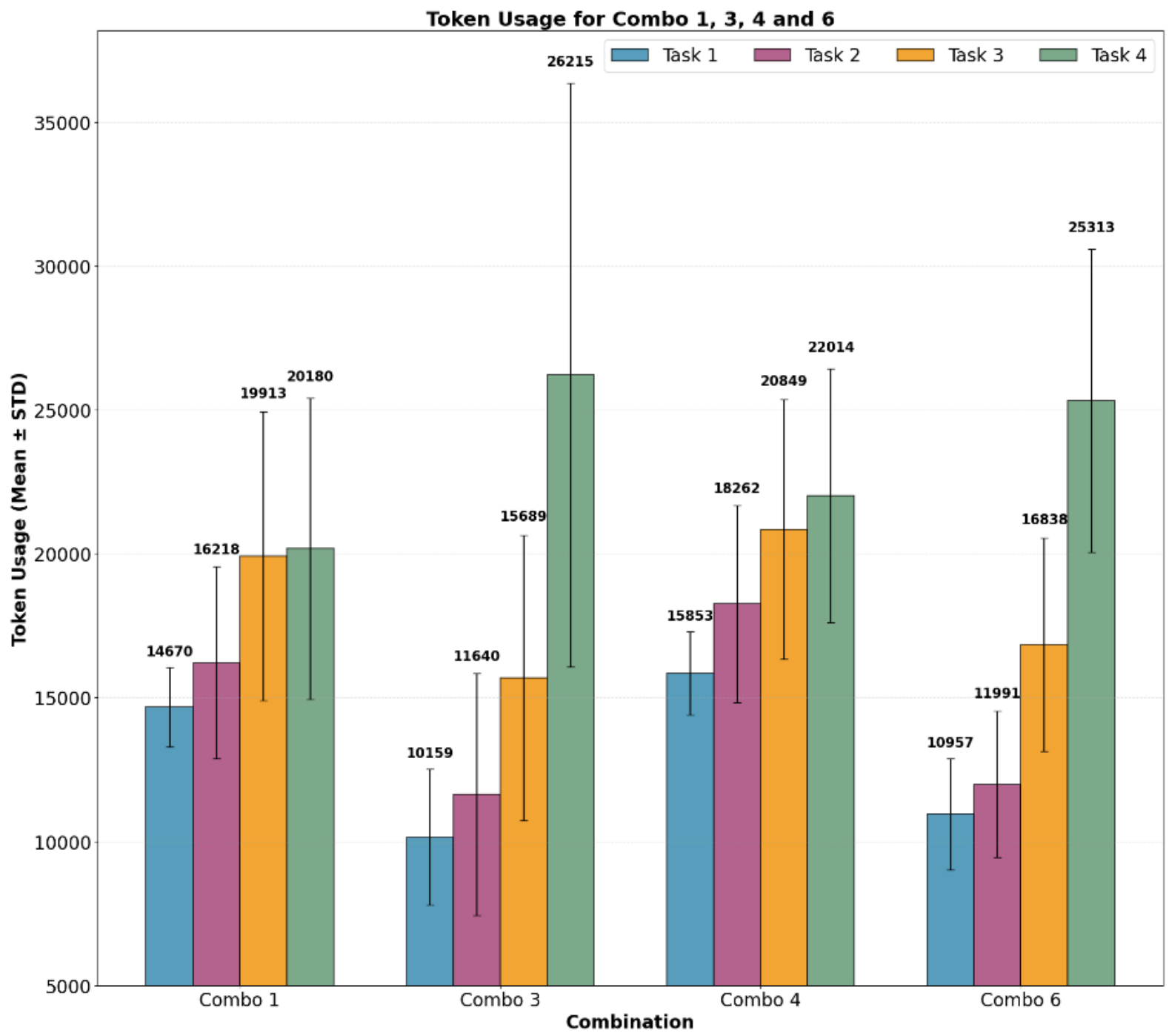

**Figure 8.** Token usage for Combo 1, 3, 4 and 6 across the four tasks

Interaction analysis revealed that *Critic_Add* systematically replicated *Critic*'s assessments without substantive contribution. In Task 2, this redundancy was particularly detrimental when paired with *Executor*. Among Combo 6's 36 failures, all due to boundary-condition errors, with no execution failures, consistent with *Critic_Add* replicating *Critic* rather than performing an independent check.

Examination of modification patterns clarified how the *Executor* amplified the negative effect of redundancy. Despite similar mean tokens (Combo 3: 11640; Combo 6: 11991), outcomes diverged. Conditional on execution success, Combo 3 achieved 6 task successes (15.8%), versus 4 (10%) for Combo 6. A recurring sequence: *Executor: "code runs"; Critic: "looks good"; Critic_Add: "looks good"*, created a triple-layer validation loop that accelerated premature consensus. The result was executable yet physically incorrect code that agents jointly accepted as final solution.

This detrimental effect arises from the interaction of three mechanisms: (1) information redundancy, where identical agent types provide overlapping assessments; (2) conversation bandwidth limitations, where redundant responses consume valuable turns without adding insights; and (3) premature consensus, where repeated agreement creates an illusion of rigorous validation. Together, these mechanisms explain why adding same-type agents reduces performance.

These findings reveal a critical principle: in turn-limited workflows, multi-round iterative inspection outperforms single-round rapid consensus. Adding a redundant reviewer (*Critic_Add*) who affirms existing assessments prematurely ends discussion, consuming valuable conversational capacity without contributing deeper analysis. This mirrors the *"groupthink"* phenomenon discussed in Section 3.2, where harmony is prioritized over critical evaluation.

The results also align with the "*functional diversity*" theory by Hong and Page (2004), which posits that cognitive diversity contributes more to collective problem-solving than individual ability. In this context, the *Executor* and *Critic* provide functionally diverse validation (runtime versus specification-based), whereas *Critic* and *Critic_Add* are functionally redundant. This redundancy leads to what Steiner (1972) describes as "*process loss*" in group productivity, where additional members impose coordination costs that outweigh potential benefits.

Furthermore, successful cases involved more discussion than failed ones (Combo 3 successes: 13,172 vs failures: 11,370; Combo 6 successes: 14,604 vs failures: 11,701). This pattern reflects what Stasser and Titus (1985) describe as "*information sampling advantage"*: groups that exchange and integrate unshared information achieve better outcomes than those that merely reiterate shared knowledge. In our system, Redundant agents reinforced shared errors, leading to less diagnostic exchanges and suboptimal consensus.

### 3.4 Verification–Validation Gap in Physical Correctness

Building on the structural and behavioural limitations identified earlier, we examined whether similar misalignments persist at the level of physical validation. Across Tasks 3 and 4, no agent combination passed physical validation, even though collaborative configurations improved code execution.

Among all combinations, only Combo 3 (*Coder* + *Executor* + *Critic*) produced any visually successful cases in Tasks 3 and 4 (two per task, resembling Figure 1c and d), but these did not pass physical validation. However, following the validation methodology outlined by Ni and Buehler (2024), visual similarity alone does not constitute verification; physical correctness requires inspection of constitutive equations and material parameters in the generated code.

Closer inspection of these seemingly successful cases revealed fundamentally incorrect stress formulations, despite correct geometry and boundary conditions. As shown in Figure 9a, the model applied the following equation for 2D plane stress analysis:

*Sigma (u) = 2*E*nu/(1-nu**2)*epsilon(u) + E/(1+nu)*tr(epsilon(u))*Identity(2)*

In contrast, the correct 2D plane stress formulation (Zienkiewicz & Cheung, 1967) is:

*Sigma (u) = (E/(1 - nu**2))*((1 - nu)*epsilon(u) + nu*tr(epsilon(u))*Identity(2))*

As illustrated in Figure 9b, the reference implementation (Ni & Buehler, 2024) employed Hooke's law with appropriate simplifications. While a 3D form can be adapted to 2D plane-stress settings, the agents produced a hybrid derivation that matched neither standard, indicating a failure to represent the underlying physical relations.

```
from fenics import *
from mshr import *
import matplotlib.pyplot as plt

# Define material properties
E = 1.0e9  # Young's modulus in Pa
nu = 0.3  # Poisson ratio

# Create mesh with a hole
domain = Rectangle(Point(0, 0), Point(1, 1)) - Circle(Point(0.5, 0.5), 0.2)
mesh = generate_mesh(domain, 50)

# Define function space
V = VectorFunctionSpace(mesh, 'P', 1)

# Define boundary conditions
def right_boundary(x, on_boundary):
    return near(x[0], 1.0) and on_boundary

def left_boundary(x, on_boundary):
    return near(x[0], 0.0) and on_boundary

bc_right = DirichletBC(V, Constant((0.0, 0.1)), right_boundary)
bc_left = DirichletBC(V, Constant((0.0, 0.0)), left_boundary)
bcs = [bc_right, bc_left]

# Define strain and stress
def epsilon(u):
    return 0.5*(nabla_grad(u) + nabla_grad(u).T)

def sigma(u):
    return 2*E*nu/(1-nu**2)*epsilon(u) + E/(1+nu)*tr(epsilon(u))*Identity(2)

# Define variational problem
u = TrialFunction(V)
v = TestFunction(V)
f = Constant((0, 0))
a = inner(sigma(u), epsilon(v))*dx
L = dot(f, v)*dx

# Solve variational problem
u = Function(V)
solve(a == L, u, bcs)

# Save displacement to file
u.rename("displacement", "label")
file = File("displacement_shear_hole.pvd")
file << u

# Plot displacement
plot(u)
plt.savefig('3.png')
plt.show()
```

```
# filename: displacement_shear_hole.py

from fenics import *
from mshr import *
import matplotlib.pyplot as plt

# Define domain and create mesh
domain = Rectangle(Point(0, 0), Point(1, 1)) - Circle(Point(0.5, 0.5), 0.2)
mesh = generate_mesh(domain, 50)
V = VectorFunctionSpace(mesh, 'P', 1)

# Define boundary condition
def left_boundary(x, on_boundary):
 return on_boundary and near(x[0], 0)
def right_boundary(x, on_boundary):
 return on_boundary and near(x[0], 1)
bc_left = DirichletBC(V, Constant((0, 0)), left_boundary)
bc_right = DirichletBC(V, Constant((0, 0.1)), right_boundary)
bcs = [bc_left, bc_right]

# Define strain and stress
def epsilon(u):
 return 0.5*(nabla_grad(u) + nabla_grad(u).T)

E = 1e9 # Young's modulus in Pa
nu = 0.3 # Poisson's ratio
mu = E / (2 * (1 + nu)) # Shear modulus (Lame's first parameter)
lambda_ = E * nu / ((1 + nu) * (1 - 2 * nu)) # Lame's second parameter

def sigma(u):
 return lambda_*div(u)*Identity(d) + 2*mu*epsilon(u)

# Define variational problem
u = TrialFunction(V)
d = u.geometric_dimension() # space dimension
v = TestFunction(V)
f = Constant((0, 0))
T = Constant((0, 0))
a = inner(sigma(u), epsilon(v))*dx
L = dot(f, v)*dx + dot(T, v)*ds

# Compute solution
u = Function(V)
solve(a == L, u, bcs)

# Save solution to file in VTK format
vtkfile = File('displacement_shear_hole.pvd')
vtkfile << u

# Plot solution
plot(u, title='Displacement', mode='displacement')

# Save plot to PNG file
plt.savefig("displacement_shear_hole.png")
```

**Figure 9. a)** Combo 3 outputs and **b)** reference Hooke's-law implementation (Ni & Buehler, 2024)

These findings expose a critical limitation: no combination verified constitutive-equation correctness, despite multiple *Critic/Rebuttal* interventions. Visually correct outputs still contained physically invalid stress formulations that went undetected throughout the dialogue.

This systematic failure exemplifies the verification–validation gap (Oberkampf & Trucano, 2000): programs may pass verification (execute correctly) yet fail validation against physical reality. The agents' failure to detect physically inconsistent formulations aligns with model inadequacy concerns; computational models necessarily approximate reality and may omit critical relations

From a knowledge-representation standpoint (Davis et al., 1993), models act as surrogates that inevitably omit aspects of reality. This perspective helps explain why language models excel at syntactic pattern matching yet may lack the domain-level representations required for physical correctness. The consistent failure of all agent combinations to detect constitutive-equation errors highlights a persistent challenge in computational validation: distinguishing acceptable simplifications from fundamental physical errors, long noted in the modelling-and-simulation literature.

Taken together, these results extend Section 3.1: multi-agent LLMs can handle geometry and boundary conditions in isolated cases but remain unreliable for verifying physical correctness. Their outputs often appear plausible yet conceal substantive physical errors, potentially leading to misleading engineering conclusions if used without expert oversight. Parallel reports (Gao et al., 2024; Zhou et al., 2024) describe similar patterns in other LLM-agent settings, high execution success masking semantic or scientific inaccuracies. These parallels indicate that the classical verification–validation gap remains unresolved in contemporary AI, re-emerging under new computational paradigms.

Although this study focuses on FEA, the mechanisms, systematic affirmation (Section 3.2) and the verification-validation gap, may recur in domains that require domain-specific rather than purely syntactic fidelity (e.g., theorem proving, legal argument construction, clinical reasoning), while more advanced models might improve domain reasoning, the interaction dynamics of role conflict, conformity, and consensus formation are likely to persist as properties of the architecture, not merely model scale; greater social-reasoning capacity could even intensify consensus-seeking behaviours

## 4. Limitations

This study has several limitations. First, scope. We focus on linear-elastic FEA and four canonical tasks; results may not generalize to nonlinear materials, contact, or transient dynamics. Second, orchestration budget. All trials used a fixed 12-turn budget; conclusions about interaction dynamics are therefore conditioned on turn-limited workflows. Third, execution pathway. For combinations without an *Executor*, code was post-hoc executed after the dialogue, whereas *Executor-inclusive* setups received inline feedback; this design choice may introduce pathway differences in how errors are surfaced. Fourth, model homogeneity. Agents were instantiated from the same base model; findings about affirmation and premature consensus may differ under heterogeneous models or toolsets. Fifth, statistical reporting. We report exact *p*-values using Fisher's exact/Chi-squared tests for binary outcomes and ANOVA for token counts; analyses are exploratory across multiple tasks and combinations, and *p*-values are reported without multiplicity adjustment. Sixth, verification levels. Our evaluation separates execution, visual/specificationification, and physical checks; the physical validation used constitutive-equation and parameter auditing in Tasks 3–4 and does not include comparison to experimental data. Seventh, reproducibility. Temperature was set to 0, but decoding remains stochastic at the token level; results reflect distributions over 40 trials per combination–task and may vary with prompts, seeds, or tool versions.

These constraints clarify the context of our findings: the evidence supports claims about role complementarity vs redundancy, systematic affirmation, and the verification–validation gap within the stated setting; extending these claims to other physics regimes, agent heterogeneity, or longer interaction budgets remains future work.

## 5. Conclusion

In this study, we dissected how role composition and interaction patterns shape the reliability of LLM-based multi-agent systems for linear-elastic FEA. Across 1,120 trials, performance depended more on functional complementarity than on team size: configurations that combined distinct roles (e.g., *Coder–Executor–Critic*) outperformed larger setups with redundant reviewers. At the interaction level, we observed systematic affirmation, *Rebuttal* agents frequently endorsed the *Critic*, including when diagnoses were wrong, and premature consensus amplified by reviewer redundancy. While the *Executor* improved execution success, this did not translate into higher task or physical-validation success, clarifying a persistent misalignment between what runs and what is right.

At the validation level, Tasks 3–4 showed a robust verification–validation gap: visually plausible outputs often encoded incorrect constitutive relations or material parameters, and no agent combination passed physical validation. These findings motivate operational design principles: (i) prioritize complementary roles over agent count; (ii) enforce multi-level checks that separate execution, visual/specification compliance, and physical correctness; and (iii) structure dialogues to delay consensus (e.g., error-triggered adversarial turns, redundancy caps). Within our 12-turn, homogeneous-model setting, these results delineate where current multi-agent orchestration helps—and where it fails. Future work should test heterogeneous agent pools, extend to nonlinear/contact regimes, and formalize protocol steps (constitutive-audit and parameter-audit turns) that raise the floor on physical fidelity

## Declaration of generative AI and AI-assisted technologies in the writing process

During the preparation of this work the authors used GPT 5 and Claude 3.7 Sonnet in order to improve writing grammar and fluency. After using this tool, the authors reviewed and edited the content as needed and took full responsibility for the content of the publication.

## Appendix A: Agent Profiles and Task Descriptions

**Table A1. Agents Profile** (Ni & Buehler, 2024)

| Agent | Agent profile |
|---|---|
| **Coder** | Engineer. You follow the query to generate codes. You write Python code to solve tasks. Wrap the code in a code block that specifies the script type. The user can't modify your code. So do not suggest incomplete code which requires others to modify. Don't use a code block if it's not intended to be executed by the executor. Don't include multiple code blocks in one response. Do not ask others to copy and paste the result. If the result indicates there is an error, fix the error and output the code again. Suggest the full code instead of partial code or code changes. If the error can't be fixed or if the task is not solved even after the code is executed successfully, analyse the problem, revisit your assumption, collect additional info you need, and think of a different approach to try. Regenerate the code whenever an expert agent makes suggestions (Ni & Buehler, 2024). |
| **Executor** | *A local Python executor feeds the full execution results into the workflow.* |
| **Critic** | You are an FEniCS expert, you make sure Coder strictly program based on the given software format, you discover the potential error and provide suggestions to Engineer. Do not generate any code. |
| **Rebuttal** | You always think differently from Critic, discuss with Critic to generate a common solution based on the query and code. Do not generate any code. |

**Table A2. Task Description** (Ni & Buehler, 2024)

| **Query Task** | Task Prompt |
|---|---|
| Task 1<br>Displacement calculation | A 2D plate occupies 1m-by-1m domain.<br>It is assumed as linear elastic and has Young's modulus of 1GPa and Poisson ratio of 0.3.<br>There is a 0.1m displacement applied on the right edge.<br>The left edge is fixed.<br>The top and bottom edges are free to move.<br>Check your formula online if you need to.<br>Define your variables.<br>Solve for the displacement using FEniCS with a mesh of 50x50, and plot the displacement |
| | result in a PNG file named 1.png |
| Task 2<br>Shear displacement calculation | Let's change the boundary condition on the right edge to a shear case.<br>The displacement along y direction is 0.1m on the right edge.<br>Define your variables.<br>Please refine the mesh to 50-by-50 elements, solve the problem again and save result into another png file |
| Task 3<br>Shear displacement calculation w/ circular hole | Let's add a circular hole of radius 0.2m in the middle of the original square domain.<br>Define your variables.<br>Please solve the shear problem and plot results |
| Task 4<br>Stress component sigma xy distribution | Let's also calculate the stress component sigma_xy and save it into another png file |

## Appendix B: Evaluation Protocol

We applied a three-level evaluation protocol: (i) code execution (verification); (ii) specification/visual compliance (match to Figure 1 references); and (iii) physical correctness (validation) for Tasks 3–4. We report code success rate, task success rate, physical-validation pass/fail, and token usage per run.

For success rates, we report binomial standard error (SE) using:

$$SE = \sqrt{\frac{p(1-p)}{n}}$$

where $\boldsymbol{p}$ denotes the observed success proportion and $\boldsymbol{n}$ is the number of trials (n=40 per combination–task).

For token usage, the standard deviation (SD) was computed as:

$$SD = \sqrt{\frac{\sum_{i=1}^{n}(x_i - \bar{x})^2}{n-1}}$$

where $\boldsymbol{x_i}$ represents individual token counts, $\boldsymbol{\bar{x}}$ is the mean token usage, and $\boldsymbol{n}$ is the number of trials.

We used Fisher's exact/Chi-squared tests for binary outcomes (success rates) and one-way ANOVA for continuous token usage; *p-values* are reported with corresponding effect sizes (e.g., Cramér's V for binaries; $\eta^2$ for ANOVA).

# Appendix C: Task Error Distribution Analysis

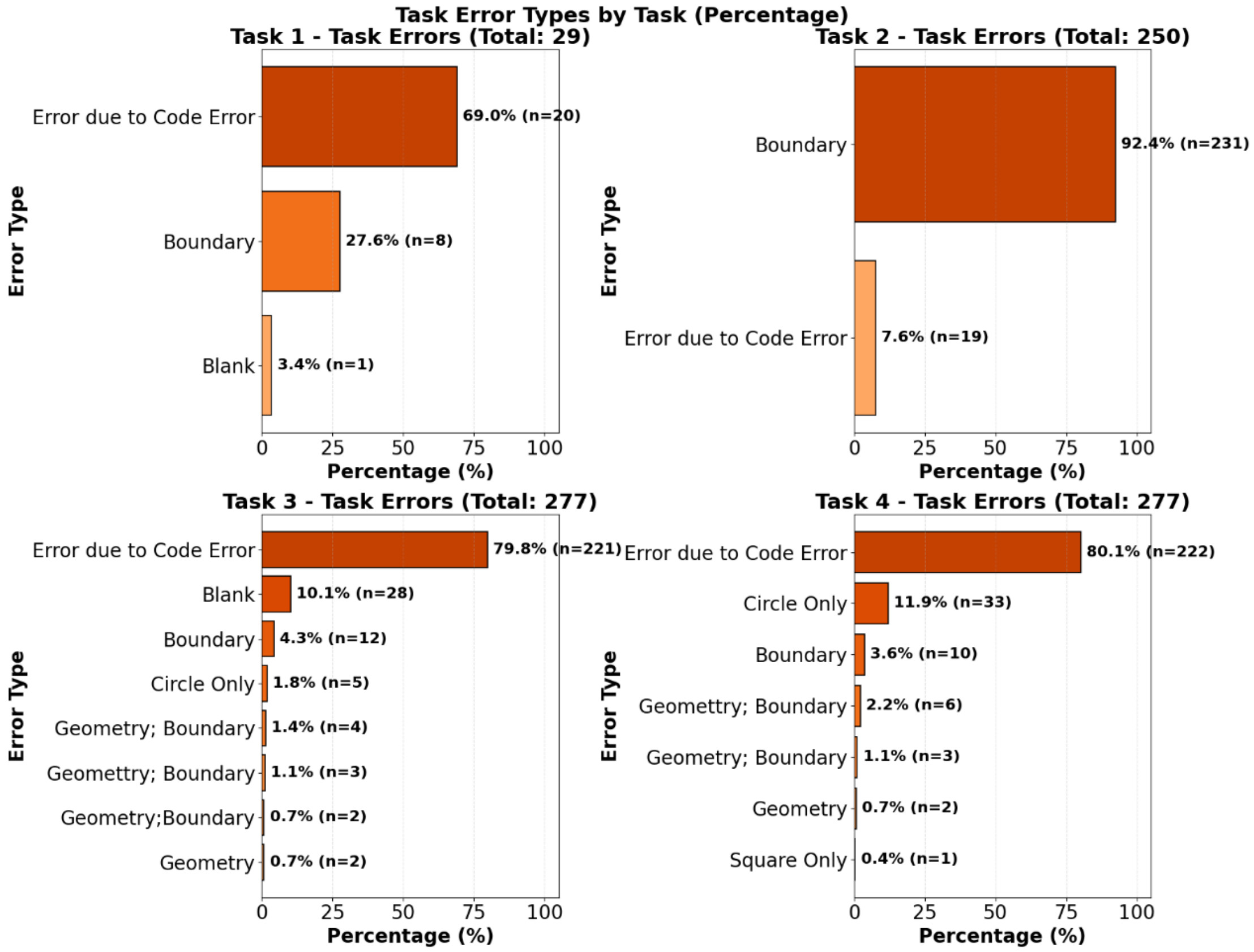

**Figure C1.** Task error distribution of all combinations

Figure C1 summarizes error-type distributions across the four tasks and complements Section 3.1 on the *Executor's* role in fixing code errors. Chi-squared tests assessed differences in error-type distributions across tasks. The overall test indicated strong differences ($p < 0.001$; *Cramér's* $V = 0.53$), suggesting that patterns vary systematically by task rather than reflecting random variation.

Pairwise comparisons showed Task 2 has a distinct error profile relative to the other tasks, with the strongest contrasts versus Task 3 and Task 4 (both $p < 0.001$). These results are consistent with Section 3.1: agents often fixed code errors yet introduced boundary-condition issues. The effect was most pronounced in Task 2, where boundary errors accounted for 92.4% of all failures despite robust initial code generation.

Task 1 was dominated by code errors (68.6%), which agents generally fixed. This aligns with Wen et al. (2024), who distinguish error categories that are fixable versus those requiring re-implementation, matching our observation that most Task-1 errors were resolved via simple corrections. By contrast, Tasks 3 and 4 showed higher proportions of code errors (79.8% and 80.1%) that were more resistant to correction. This pattern is consistent with increasing algorithmic complexity from Task 1 to Task 4, which posed progressively greater challenges for the agents. Although Tasks 3 and 4 appear visually similar in their aggregated distributions, statistical testing still found significant differences between them ($p < 0.001$), suggesting that subtle requirement differences can yield distinct error patterns even in visually comparable problems, highlighting the task-specific nature of agent performance.

The boundary-error pattern in Task 2 parallels Khan et al. (2025), who report difficulties with boundary-relationship identification in visualization tasks. This convergence suggests boundary-related challenges may be intrinsic to certain visualization task types rather than solely a limitation of current models.

Our observed link between task complexity and error patterns is also consistent with Wang et al. (2025), who show that complexity influences both error types and correction difficulty: across 557 incorrect code samples, different tasks consistently triggered distinct error patterns across model architectures, pointing to fundamental interactions between task characteristics and AI reasoning strategies.

Collectively, these findings indicate that task-dependent error patterns reflect systematic interactions between task characteristics and current AI methods. Future work should move beyond general model improvements toward task-tailored error-handling mechanisms, especially for boundary-focused visualization tasks.